\documentclass[10pt,twocolumn,letterpaper]{article}

\usepackage{cvpr}              % To produce the CAMERA-READY version
\definecolor{cvprblue}{rgb}{0.21,0.49,0.74}
\usepackage[pagebackref,breaklinks,colorlinks,allcolors=cvprblue]{hyperref}

\def\paperID{55} % *** Enter the Paper ID here
\def\confName{CVPR}
\def\confYear{2026}

\title{A Two-Stage Dual-Modality Model for Facial Expression Recognition}
\begin{document}
\author{
	Jiajun Sun \quad Zhe Gao$^{\dagger}$ \\
	Shanghai Normal University \\
	{\tt\small \{ora942878@gmail.com, zgao0911@shnu.edu.cn\}} 
}
\maketitle

\begin{abstract}
	
This paper addresses the EXPR challenge in the 10th Affective Behavior Analysis in-the-Wild (ABAW) workshop and competition, which requires frame-level classification of eight expression categories from unconstrained videos. These unconstrained conditions introduce substantial variation and noise across adjacent frames in raw videos, making accurate frame-level EXPR recognition difficult.

We propose a two-stage dual-modal (audio-visual) model to address these challenges. In Stage I, we train a DINOv2-based encoder on external image-level facial expression datasets to learn more expression-aware visual representations. During this stage, we introduce padding-aware augmentation (PadAug) to improve robustness to boundary artifacts caused by large face crops, together with a training-only mixture-of-experts (MoE) head to provide stronger task-oriented supervision.

Stage II focuses on modality fusion and temporal consistency. For the visual modality, faces are re-cropped from raw videos at multiple scales, and the resulting visual features are extracted via the DINOv2-based encoder pretrained in Stage I and averaged to form robust frame-level visual representations. In parallel, audio features are extracted from short windows of the same videos using Wav2Vec~2.0 and aligned with the visual features at the frame level to provide complementary acoustic cues. The two modalities are integrated through a lightweight gated fusion module, and inference-time temporal smoothing is further applied to improve prediction stability.

Experiments and ablation studies verify the effectiveness of the proposed method. The final system achieves a macro-averaged F1 score of 0.5368 on the official validation set, 0.5122 $\pm$ 0.0277 under 5-fold cross-validation, and 0.391 on the official challenge server. These results show that the proposed combination of visual adaptation, audio-visual fusion, and temporal smoothing is effective for robust frame-level EXPR recognition in unconstrained environments.

\end{abstract}

\section{Introduction}

Frame-level facial expression recognition (EXPR) from raw videos is an important task in affective and behavioral analysis, especially in in-the-wild settings (e.g., unconstrained environments). In recent years, EXPR has attracted increasing attention because of its broad applications in fields such as human-computer interaction, affective computing, and behavioral analysis~\cite{poria2017review,Li2022DeepFERSurvey,Kopalidis2024AdvancesFERSurvey}. To advance EXPR research, several important benchmarks have been established, notably the Aff-Wild and Aff-Wild2 datasets~\cite{kollias2019expression} and the Affective Behavior Analysis in-the-Wild (ABAW) workshop series~\cite{kollias2025advancements,kollias2025emotions}.

The 10th ABAW workshop in 2026 continues this effort~\cite{kollias2026abaw10}. In this year’s EXPR track, each frame in Aff-Wild2~\cite{kollias2019expression} is assigned to one of eight categories: Neutral, Anger, Disgust, Fear, Happiness, Sadness, Surprise, and Other. Compared with controlled laboratory settings, real-world scenarios present substantial challenges for accurate and consistent facial expression recognition. In raw videos of Aff-Wild2, for instance, face localization is often inaccurate or unstable, facial scale may vary substantially across frames, and many samples are affected by blur, occlusion, extreme pose, illumination changes, and low image quality~\cite{gera2023exprwild}. These issues make visual evidence noisy and inconsistent across time and hence make reliable emotion perception difficult~\cite{Xue_2022_CVPR,savchenko2024emotieffnet,yu2024semitemporal}.

In addition, affective signals in real-world videos are inherently in multi-modality~\cite{poria2017review}. Visual information, such as facial appearance, may be insufficient for accurate expression recognition in ambiguous cases, while audio information, such as interjections and plosive sounds, can provide critical complementary cues~\cite{zhang2022transformer,dresvyanskiy2024sun,kim2024cascaded,Savchenko_2025_CVPR}. However, effectively integrating data in multi-modality, such as visual and audio streams, remains challenging, especially when frame-level predictions fluctuate over time under unconstrained environments~\cite{arevalo2017gmu,tsai2019mult,mittal2020m3er,Xue_2022_CVPR}.

To address these challenges, we propose a two-stage dual-modality framework for robust frame-level facial expression recognition in this paper.
\pagebreak
\par\smallskip
\noindent\textbf{Stage I: Visual adaptation}\par
In Stage I, we pretrain a DINOv2-based visual encoder~\cite{oquab2023dinov2} on external facial expression datasets of RAF-DB~\cite{li2019rafdb} and AffectNet~\cite{mollahosseini2017affectnet} to learn extra expression-aware visual representations. In this stage, we also employ two specific designs to strengthen the performance: the padding-aware augmentation (PadAug) which improves the robustness of our model to face-cropping boundaries and scale variations, and a training-only MoE head that provides stronger task-oriented supervision during visual adaptation.

\par\smallskip
\noindent\textbf{Stage II: Frame-level audio–visual recognition}\par
Given raw videos, faces are re-cropped at three scales and frame-level visual features are extracted via the DINOv2 encoder adapted in Stage I. These features are then aggregated to form robust visual representations. In parallel, acoustic features are extracted from a pretrained Wav2Vec~2.0 model~\cite{baevski2020wav2vec} and aligned with the visual representations to provide a necessary complement from the audio modality. 

The two parallel features are integrated through a lightweight gated fusion module~\cite{arevalo2017gmu} with a small parameter count, followed by inference-time temporal smoothing~\cite{Xue_2022_CVPR} to provide enhanced EXPR predictions with high accuracy and stable temporal consistency.

Our proposed model is evaluated on Aff-Wild2, the official dataset in ABAW, and achieves 0.5368 Macro-F1 on the official validation set, 0.5122 $\pm$ 0.0277 Macro-F1 under 5-fold cross-validation, and 0.391 Macro-F1 on the final official challenge test set. These results consistently demonstrate the strong effectiveness and competitive strength of the proposed framework across different evaluation settings.

\noindent\textbf{Our contributions are summarized as follows:}
\begin{itemize}

	\item The dual modality framework of the proposed model provides outstanding interpretability on facial expressions from the raw videos. The majority of emotional information is captured by the visual encoder finetuned in Stage I since images usually constitute the predominant information in videos while the supplementary acoustic cues provided in Stage II put the last piece of the puzzle in place. 
	
	\item The visual understanding and representation for the visual encoder are enhanced through two complementary modules with distinct specializations. The padding-aware augmentation (PadAug) strategy improves model robustness to padding artifacts and scale variation, and a training-only Mixture-of-Experts (MoE) head strengthens visual adaptation effectiveness without increasing inference complexity.
	
	\item A lightweight frame-aligned gated audio--visual fusion module, together with an inference-time temporal smoothing module, delivers clear performance gains on our model. This design requires only a comparatively small number of additional parameters, emphasizing the effectiveness from simpler integration with low overhead.
\end{itemize}

The rest of this paper is organized as follows. Section 2 reviews related work on frame-level EXPR recognition in ABAW. Section 3 presents our proposed two-stage dual modality method. Section 4 describes the experimental settings and reports the main results and  ablation studies. Section 5 concludes the paper and discusses future work.

\section{Related Work}

Frame-level facial expression recognition (EXPR) in-the-wild aims to assign an expression label to each frame in unconstrained videos. Compared with lab-controlled settings, this task must handle large variations in pose, illumination, occlusion, motion blur, identity, and annotation ambiguity. The ABAW benchmark series, built on Aff-Wild2~\cite{kollias2019expression}, has become one of the main evaluation platforms for this problem. Across successive editions of ABAW~\cite{kollias2025advancements,kollias2025emotions}, EXPR has generally been studied together with other affective behavior tasks rather than as a standalone setting, including valence--arousal estimation, action unit (AU) detection, emotional reaction intensity estimation, compound expression recognition, and related multi-task formulations~\cite{kollias2019face,kollias2021distribution,kollias2023multi}. More broadly, recent surveys show that EXPR has become a core topic in affective computing and facial behavior analysis~\cite{poria2017review,Li2022DeepFERSurvey,Kopalidis2024AdvancesFERSurvey,gera2023exprwild}.

Different datasets and modalities have been adopted in EXPR research. Aff-Wild2 remains the dominant benchmark in ABAW-related work. External facial datasets such as AffectNet~\cite{mollahosseini2017affectnet}, RAF-DB~\cite{li2019rafdb}, CK+~\cite{LuceyCKPlus}, FER+~\cite{barsoum2016training}, AffectNet+~\cite{11259096}, and C-EXPR~\cite{kollias2023multi} are frequently used for pretraining, transfer learning, or auxiliary supervision. In terms of modality, most EXPR systems remain visual-only~\cite{savchenko2022mobileexpr,xue2023exprssl,lin2024robustclip,ma2023maeface,savchenko2024emotieffnet,yu2024semitemporal}, while other methods incorporate audio-visual fusion~\cite{zhang2022transformer,liu2023multimodalexpr,dresvyanskiy2024sun,kim2024cascaded} or extend further to audio-visual-text settings~\cite{wang2024affective,Savchenko_2025_CVPR}.

At the model level, visual EXPR methods mainly rely on either CNN backbones or Transformer-style encoders. The representative of CNN-based directions include ResNet~\cite{he2016resnet}, EfficientNet~\cite{tan2019efficientnet}, and task-oriented efficient variants such as EmotiEffNet~\cite{savchenko2024emotieffnet}. The representative of Transformer-based directions include ViT~\cite{dosovitskiy2021image}, Swin Transformer~\cite{liu2021swin}, CLIP~\cite{radford2021learning}, MAE~\cite{he2022mae}, and DINOv2~\cite{oquab2023dinov2}. In affective behavior analysis and facial expression recognition, these foundation-style visual encoders have already been adapted to downstream tasks, including CLIP-based EXPR systems~\cite{lin2024robustclip}, MAE-based EXPR systems~\cite{ma2023maeface}, DINOv2-based ABAW competition systems~\cite{li2024taskadaptive}, and recent DINO-prior FER models~\cite{xie2025dinoprior,wang2025dinoprior}. Performance gains are often associated with coarse-to-fine prediction~\cite{Xue_2022_CVPR}, self- or semi-supervised pretraining~\cite{xue2023exprssl,yu2024semitemporal,he2022mae,oquab2023dinov2}, cross-attention-based interaction~\cite{zhang2022facial,kim2024cascaded}, AU-aware or multi-task supervision~\cite{kollias2019face,kollias2021affect,kollias2021distribution,kollias2024distribution}, and model ensembling~\cite{dresvyanskiy2024sun,Savchenko_2025_CVPR}.

Regarding other modalities in EXPR, audio streams are commonly encoded either by handcrafted acoustic descriptors or by pretrained speech representation models such as wav2vec~2.0~\cite{baevski2020wav2vec} and HuBERT~\cite{hsu2021hubert}; these types of encoders have been used in ABAW submissions for audio-visual emotion understanding~\cite{zhang2022transformer,dresvyanskiy2024sun,Savchenko_2025_CVPR}. Textual branches typically rely on BERT~\cite{devlin2019bert} or RoBERTa~\cite{liu2019roberta}, and some recent systems further use generated or prompted text as auxiliary semantic context~\cite{Savchenko_2025_CVPR,wang2024affective}. For modality fusion, prior work explores direct feature concatenation~\cite{zhang2022transformer,liu2023multimodalexpr}, gated fusion~\cite{arevalo2017gmu}, tensor fusion~\cite{zadeh2017tensor}, cross-modal attention or multimodal Transformers~\cite{tsai2019mult,kim2024cascaded}, and decision-level combination or ensembling~\cite{dresvyanskiy2024sun,Savchenko_2025_CVPR}.

Because raw videos contain strong temporal dependencies, competitive EXPR systems usually add temporal modeling or temporal stabilization. Explicit temporal modeling includes LSTM-based sequence learning~\cite{hochreiter1997lstm,cabacas2025lstmddamfn}, temporal convolutional networks~\cite{Lea2017TCN,savchenko2024emotieffnet}, Transformer-based sequence encoders~\cite{vaswani2017attention,zhao2021formerdfer,ma2023logoformer}, and sliding-window prediction over local frame neighborhoods~\cite{savchenko2022mobileexpr,yu2024semitemporal,liu2023multimodalexpr}. Implicit temporal stabilization is often performed at inference time through smoothing or window-based post-processing methods~\cite{Xue_2022_CVPR,savchenko2022mobileexpr,xue2023exprssl,ma2023maeface,savchenko2024emotieffnet}. Overall, recent competitive systems tend to combine strong visual representations with lightweight multimodal fusion and stable temporal aggregation~\cite{dresvyanskiy2024sun,Savchenko_2025_CVPR,yu2024semitemporal}.
\section{Method}

\begin{figure*}[t]
	\centering
	\includegraphics[width=\textwidth]{figures/framework.jpg}
\caption{Overview of the proposed model. For the video modality, three facial crops at different scales for each target frame are extracted from the raw video and encoded by an adapted DINOv2 backbone. For the audio modality, a Wav2Vec~2.0 encoder is used to extract window-based features aligned with the visual frame. The resulting multi-scale visual representation is fused with the audio features through a gated fusion module for frame-level facial expression recognition. During inference, lightweight temporal smoothing is further applied to improve prediction stability and temporal consistency.}

	\label{fig:framework}
\end{figure*}

\begin{figure}[t]
\centering
\includegraphics[width=\linewidth]{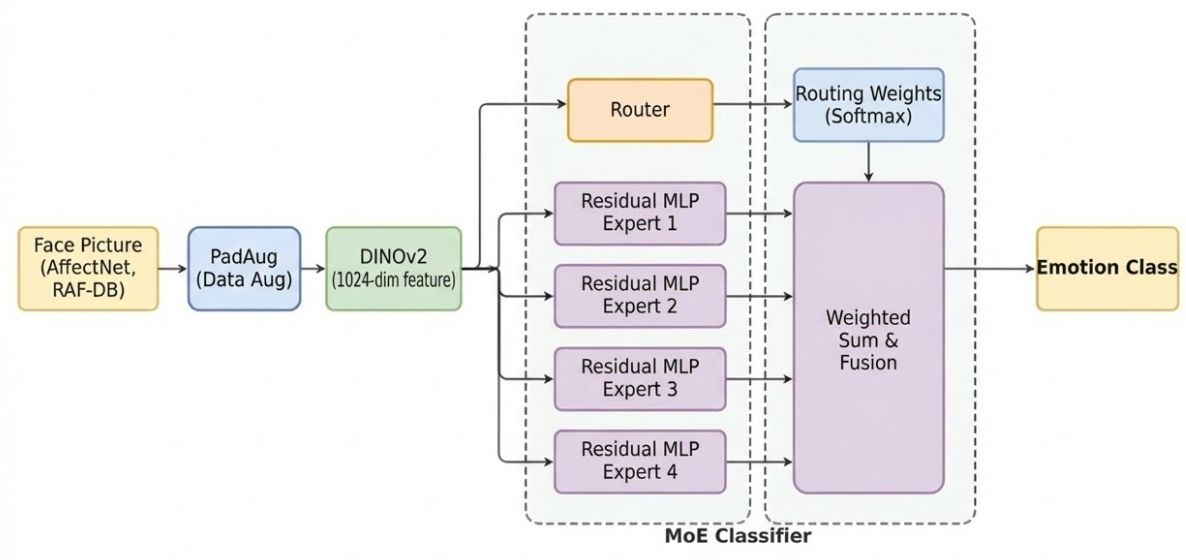}
\caption{The visual adaptation pipeline in Stage-I. The DINOv2-based encoder is adapted on AffectNet and RAF-DB. PadAug simulates boundary padding artifacts caused by large facial crops, while a training-only MoE head provides sample-dependent expert routing to enhance DINOv2 adaptation. After Stage I, the MoE head is discarded and only the adapted DINOv2 backbone is retained.
}
\label{fig:stage1_moe}
\end{figure}

\begin{figure}[t]
	\centering
	\includegraphics[width=\linewidth]{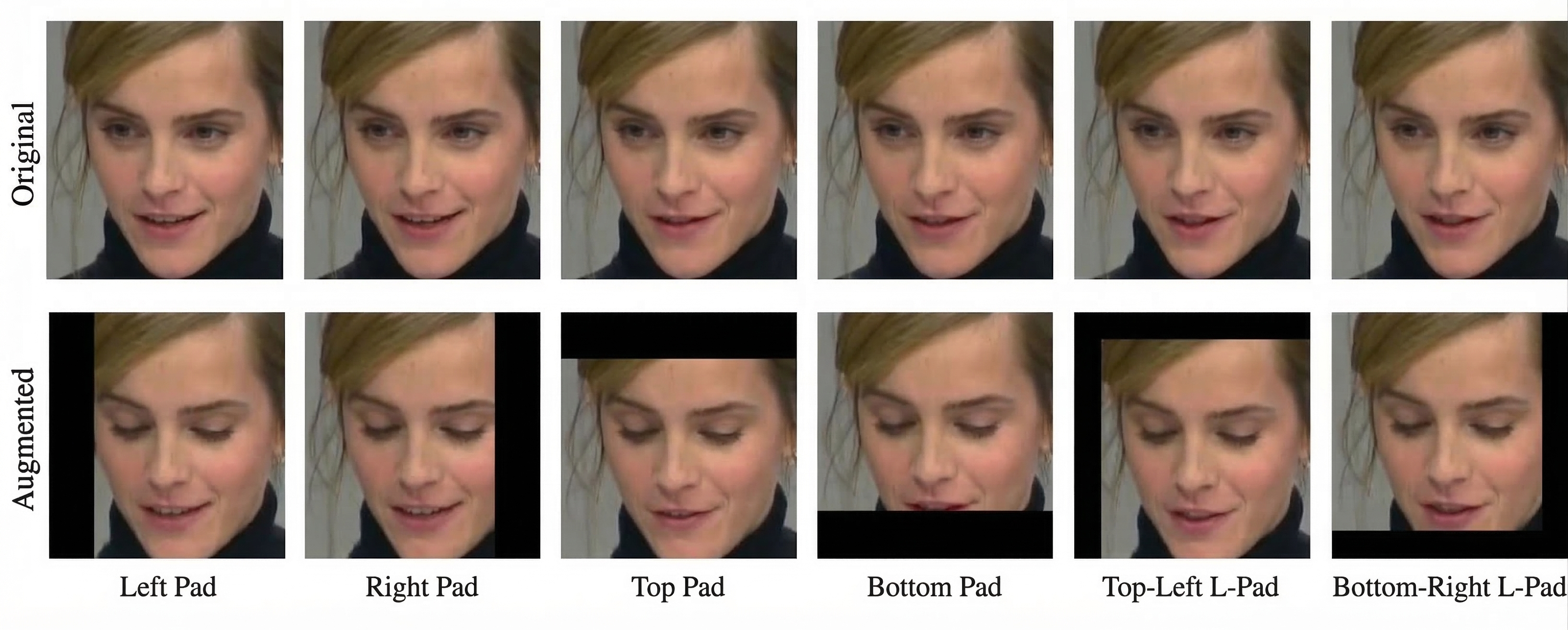}
	\caption{Illustration of the proposed padding-aware augmentation (PadAug). 
		The top row shows original facial crops, while the bottom row presents 
		augmented samples where artificial padding is applied to different image 
		boundaries to simulate boundary artifacts caused by imperfect face cropping 
		in raw videos.}
	\label{fig:padaug}
\end{figure}

Illustrated in Fig.~\ref{fig:framework}, our proposed model performs frame-level audio–visual expression recognition from raw videos in a two-stage manner. Our model re-crops the human faces directly on frame-level images from raw videos and constructs three facial crops with different scales, which are encoded by a DINOv2-based visual encoder to obtain robust multi-scale visual representations. In parallel, a short audio segment centered at the target frame from the same video is encoded by Wav2Vec~2.0 to produce the frame-aligned acoustic cues as necessary supplements. Features from the dual modalities (audio-visual) are then integrated in a lightweight gated fusion module and selected for frame-level expression classification. A simple temporal smoothing module is applied during inference to enhance prediction stability and temporal consistency. The two critical parts in this method are: visual adaptation by the finetuned DINOv2-based encoder in Stage I and the modality alignment, integration and inference via gated fusion and temporal smoothing modules.

\subsection{Stage-I Visual Adaptation}
\label{sec:stage1}
A pipeline, displayed in Fig.~\ref{fig:stage1_moe}, is designed for visual adaptation in Stage I. A pretrained model of DINOv2 ViT L/14 was selected as the encoder backbone. Two complementary components, PadAug and a training-only MoE head, are added before and after the DINOv2 backbone to enhance model robustness to padding artifacts and scale variations, thereby improving visual adaptation effectiveness without increasing inference complexity. After adaptation, the MoE head is discarded and only the finetuned DINOv2 backbone is retained for Stage-II dual modality learning on raw videos.

To improve the expression sensitivity of the visual  encoder before multimodal learning, the DINOv2 backbone was adapted on two image-level FER datasets, AffectNet~\cite{mollahosseini2017affectnet} and RAF-DB~\cite{li2019rafdb}. These datasets provide large-scale supervision for learning discriminative expression features under diverse facial appearances. Since the final target task is frame-level EXPR recognition in unconstrained videos, this intermediate adaptation serves as an efficient bridge between generic self-supervised visual pretraining and video-based expression analysis.
\subsubsection{Padding-aware Augmentation (PadAug)}
\label{sec:padaug}

A practical issue in frame-level visual recognition is that large facial crops may exceed the image boundary~\cite{innamorati2020learning}, especially when large cropping scales are used. In such cases, padded regions along image borders may introduce non-negligible distribution shifts between visually clean images and boundary-affected cropped faces. This may affect model performance for raw-video inference. Examples of cropped images are shown in Fig.~\ref{fig:padaug}.

To alleviate this mismatch during Stage-I visual adaptation, we introduce padding-aware augmentation (PadAug), which inserts black padding bars along image boundaries and applies small spatial perturbations to the padded region. As shown in Fig.~\ref{fig:padaug}, this augmentation simulates common boundary conditions, including left, right, top, and bottom padding, as well as corner-like artifacts at the top-left and bottom-right. Unlike generic data augmentations, PadAug is tailored to scale variation and boundary corruption in multi-scale facial representation learning, explicitly exposing the model to boundary patterns produced by large-scale face cropping in unconstrained videos.

\subsubsection{The MoE-assisted DINOv2 Adaptation}
\label{sec:moe_dino}

Given a face image $\mathbf{x}$, the selected visual encoder, DINOv2 ViT-L/14, produces a CLS-token feature $\mathbf{u} \in \mathbb{R}^{1024}$, which is defined as
\begin{equation}
	\mathbf{u} = E_v(\mathbf{x}),
	\label{eq:stage1_u}
\end{equation}
where $E_v(\cdot)$ denotes the encoding process of DINOv2.

To provide stronger task-oriented supervision during visual adaptation, a mixture-of-experts (MoE) classifier is attached to the DINOv2 output feature $\mathbf{u}$. The MoE head consists of a router and $M$ expert branches. The normalized feature $\mathrm{LN}(\mathbf{u})$ is fed into the router to generate the sample-dependent routing weights $\boldsymbol{\alpha} \in \mathbb{R}^{M}$ over the experts:
\begin{equation}
	\boldsymbol{\alpha} = \mathrm{Softmax}\big(R(\mathrm{LN}(\mathbf{u}))\big),
	\label{eq:stage1_alpha}
\end{equation}
where $R(\cdot)$ denotes the routing function. Each MoE expert is implemented as an MLP-based transformation. Let $\mathbf{e}_m$ denote the output of the $m$-th expert, where $m = 1, \dots, M$. The expert outputs are aggregated by a weighted summation:
\begin{equation}
	\tilde{\mathbf{u}} = \sum_{m=1}^{M} \alpha_m \mathbf{e}_m.
	\label{eq:stage1_moe}
\end{equation}

Then, the expression recognition logits $\boldsymbol{\ell}$ are produced by a normalized dropout linear classification layer:
\begin{equation}
	\boldsymbol{\ell} = W_c \, \mathrm{Drop}\big(\mathrm{LN}(\tilde{\mathbf{u}})\big) + \mathbf{b}_c,
	\label{eq:stage1_logit}
\end{equation}
where $W_c$ and $\mathbf{b}_c$ denote the classifier parameters.

Compared with a standard single-branch classifier, the MoE head allows different experts to specialize in different expression patterns and provides richer supervisory signals for adapting the visual backbone. It is worth noting that, the MoE head is used only in Stage~I. After adaptation, it is discarded, and only the adapted DINOv2 backbone is retained for visual feature extraction in Stage~II.

\subsubsection{Objective Function in Stage I}

The class-weighted cross-entropy loss is adopted to optimize the parameters in Stage~I. The objective function is formulated as
\begin{equation}
	\mathcal{L}_{\text{stage1}} = - \sum_{c=1}^{C} w_c \, y_c \log p_c,
	\label{eq:stage1_loss}
\end{equation}
where $p_c$ denotes the predicted probability for class $c$, $w_c$ is the corresponding class weight, $y_c$ is the soft target distribution, and $C=8$ is the number of expression categories.

\subsection{Stage-II: Frame-level Audio-Visual Integration}

After visual adaptation in Stage I, frame-level audio-visual modalities are integrated in Stage II for EXPR decision. In Stage II, the dual-modal (audio-visual) features are first extracted from raw videos via the pretrained Wav2Vec~2.0~\cite{baevski2020wav2vec} and the visual encoder adapted in Stage I respectively. The features from the two modalities can mutually provide complementary information for the subsequent emotional recognition. Next, the audio-visual features are aligned at frame-level, integrated and selected via a gated fusion module, then stabilized in temporal sequence through an inference-time temporal smoothing module, and finally yield EXPR results with high accuracy.

\subsubsection{Visual Representation in Stage II}

Since the official ABAW face crops are usually in relatively low-resolution and may weaken the subtle expression cues after resizing, we re-crop the facial images from the raw video frames by the buffalo\_l package under InsightFace~\cite{insightface2025,Deng2020RetinaFace}. For each frame $t$, faces are detected and three square crops centered on the detected bounding box are generated with scale factors of 0.9, 1.2, and 1.5. The generated three-square crops are then fed into the adapted DINOv2 backbone, and the extracted features are averaged to form the multi-scale visual representation. Let $x_t^{s}$ denote the crop at scale $s$ for frame $t$, the extracted visual feature $f_t^{v,s}$ from the adapted DINOv2 backbone under this scale is calculated by:
\begin{equation}
	f_t^{v,s} = E_v \left( x_t^{s} \right).
\end{equation}
and the frame-level visual representation is obtained as:
\begin{equation}
	f_t^v = \frac{1}{3}\sum_{s=\{0.9,1.2,1.5\}} f_t^{v,s}.
\end{equation}

In multi-person frames, the largest detected face is used by default. If a crop exceeds the image boundary, zero padding is applied. This strategy captures complementary information from different crop extents while keeping the visual branch simple at the aggregation stage.

\subsubsection{Frame-aligned Audio Features}

Although facial images usually contain the majority expression information from raw videos, acoustic signals, e.g., interjection and plosive sounds, could provide critical supplementary cues for EXPR, therefore, audio is adopted as the second modality in our method. To provide necessary affective cues from speech, we extract the acoustic features using a pretrained model of Wav2Vec~2.0 and hence formulate our method into a dual modality framework.

In this framework, the audio and video streams exhibit distinct temporal resolutions, consequently, frame-level alignment is required. For each target frame $t$, a short audio segment centered around this frame is used and the corresponding acoustic features are aggregated to form a frame-level audio representation. Let $z_{\tau}$ denote the acoustic feature at temporal index $\tau$ and $\Omega_t$ denote the set of indices aligned with frame $t$, the aligned audio feature can be obtained by:
\begin{equation}
	f_t^a = \frac{1}{|\Omega_t|}\sum_{\tau \in \Omega_t} z_{\tau},
\end{equation}

In our two-stage method, we use a centered temporal window of 0.50s for this aggregation. This short-window averaging provides more stable acoustic cues than direct nearest-neighbor assignment.

\subsubsection{Gated Fusion for Dual-Modal Integration}

The aligned visual feature $f_t^v$ and audio feature $f_t^a$ are integrated through a lightweight gated fusion module. First, the features from the dual modalities are projected into a shared hidden space:
\begin{equation}
	z_t^v = P_v(f_t^v), \qquad z_t^a = P_a(f_t^a),
\end{equation}
where $P_v(\cdot)$ and $P_a(\cdot)$ are learnable linear projections for the visual and audio modality respectively. Next, a gating vector is predicted from the concatenated features:
\begin{equation}
	g_t = \sigma(G([f_t^v; f_t^a])),
\end{equation}
where $G[\cdot]$ denotes the gating network, $[\cdot]$ represents the concatenation operation, and $\sigma(\cdot)$ is the sigmoid activation. The gated fusion is then calculated as:
\begin{equation}
	h_t = g_t \odot z_t^v + (1-g_t) \odot z_t^a,
\end{equation}
where $\odot$ is the symbol of element-wise multiplication. Finally, the frame-level logits are obtained by a normalization dropout classification head:
\begin{equation}
	l_t = W_o\,Drop(LN(h_t)) + b_o,
\label{eq:stage2_logits}
\end{equation}where $W_o$ and $b_o$ are the classification parameters. The gated fusion module enables our model to adaptively control the contributions from visual and acoustic cues according to their frame-wise reliability. Therefore, this module also plays a role of feature selection to some extent.

It is worth noting that the logit $l_t$ obtained via Eq.~\eqref{eq:stage2_logits} is computed independently for individual frame $t$, without incorporating the temporal consistency. The final EXPR decisions from our two-stage model are made after the temporal sequence is smoothed.

\subsubsection{The Objective Function in Stage II}

Stage II is trained on the Aff-Wild2 EXPR labels using the frame-level representations in dual modality discussed above. Since the DINOv2-based visual encoder has already been adapted in Stage I and the acoustic encoder Wav2Vec~2.0 is pretrained, Stage II only needs to optimize the gated fusion head, which substantially reduces the training cost for our dual-modal learning.

Let $p_c$ denote the predicted probability for class $c$ at frame $t$, and $y_t$ denote the ground-truth label. The class-weighted cross-entropy loss is optimized over valid labeled frames to determine the parameters in our model:
\begin{equation}
	\mathcal{L}_{\text{stage-2}} = -\sum_{c=1}^{C} w_c\,\mathbf{1}(y_t = c)\log p_c,
\end{equation}
where $w_c$ is the weight for class $c$ and $\mathbf{1}(\cdot)$ is the indicating function, when $y_t = c$, $\mathbf{1}(y_t = c)=1$, otherwise $\mathbf{1}(y_t = c)=0$. This objective function encourages discriminative dual-modal representations while partially alleviating classification issues caused by class imbalance.

\subsubsection{Inference-time temporal smoothing}

Although an expression recognition decision at the frame-level can be made via Eq.~\eqref{eq:stage2_logits}, adjacent frames in a video usually exhibit strong temporal continuity and hence direct frame-wise predictions may fluctuate by interruptions from nearby frames. To improve consistency without introducing extra complicated learning structure, a lightweight post-hoc temporal smoothing strategy is adopted for inference. After careful comparison in the  ablation study, the median smoothing method is selected. Details of the smoothing selection are shown in Section 4.

Let $l_t \in \mathbb{R}^{C}$ denote the predicted logit vector for frame $t$ and $C = 8$ is the number of expression classes. Consider a temporal window centered at frame $t$:
\begin{equation}
	\mathcal{N}(t)=\{t-k,\cdots,t,\cdots,t+k\},
\end{equation}
where the window size is $2k+1$. In the optimized system, an odd window size of 101 frames is used. Then the smoothed logit for class $c$ is defined as:
\begin{equation}
	\tilde{l}_{t,c} = \mathrm{median}\{l_{t-k,c},\cdots,l_{t+k,c}\},
\end{equation}
and the final recognitions after temporal smoothing are:
\begin{equation}
	\hat{y} = \arg\max_{c}\tilde{l}_{t,c}.
\end{equation}

\section{Experiments and Results}
\begin{table}[t]
	\centering
	\caption{Official challenge server results on the EXPR Recognition Challenge.}
	\label{tab:challenge_results}
	\begin{tabular}{lc}
		\toprule
		Teams & Total Score \\
		\midrule
		Ours & \textbf{0.391} \\
		HSEmotion                & 0.386 \\
		USTC-IAT-United          & 0.360 \\
		IMLAB                    & 0.320 \\
		baseline                 & 0.225 \\
		\bottomrule
	\end{tabular}
\end{table}

\begin{table*}[t]
\centering
\caption{Ablation on face crop scale and visual encoder training strategy. Results are reported as F1 on the official validation set using a linear classifier on top of the extracted frozen visual features.}
\label{tab:visual_ablation}
\resizebox{\textwidth}{!}{
\begin{tabular}{lccccc}
\toprule
Training Method & Official Face Crop + Linear & FaceCrop-0.9 + Linear & FaceCrop-1.2 + Linear & FaceCrop-1.5 + Linear & Multi-scale Mean (0.9/1.2/1.5) + Linear \\
\midrule
Pretrained               & 0.3007 & 0.3199 & 0.3307 & 0.3192 & 0.3355 \\
Fine-tune + MLP          & 0.3378 & 0.3697 & 0.3773 & 0.3738 & 0.3800 \\
Fine-tune + MoE          & 0.3576 & 0.4040 & 0.4251 & 0.4223 & 0.4257 \\
Fine-tune + MoE + PadAug & 0.3525 & 0.3992 & 0.4205 & 0.4245 & \textbf{0.4344} \\
\bottomrule
\end{tabular}
}
\end{table*}

\begin{table}[t]
\centering
\caption{Ablation on audio-frame alignment strategy. Results are reported as F1 on the official validation set.}
\label{tab:audio_alignment}
\begin{tabular}{lc}
\toprule
Alignment Strategy & F1 \\
\midrule
Nearest               & 0.2423 \\
Window Mean ($0.25$\,s)   & 0.2876 \\
Window Mean ($0.50$\,s)   & \textbf{0.2901} \\
Window Mean ($0.75$\,s)   & 0.2886 \\
\bottomrule
\end{tabular}
\end{table}
\subsection{Evaluation Metric}
\label{sec:metric}

Following the ABAW EXPR challenge protocol~\cite{kollias2025advancements}, we use the official EXPR 
metric, the score of macro-averaged F1 over all expression categories as the main evaluation metric. Let $C$ denote the number of expression categories. The overall EXPR score is defined as:

\begin{equation}
P_{\mathrm{EXPR}} = \frac{1}{C}\sum_{c=1}^{C} F1^{c},
\end{equation}
where $F1^{c}$ denotes the $F1$ score of the $c$-th expression category. For each class, the $F1$ score is computed as:
\begin{equation}
F1^{c} = \frac{2 \times \mathrm{Precision}_c \times \mathrm{Recall}_c}{\mathrm{Precision}_c + \mathrm{Recall}_c},
\end{equation}
where $\mathrm{Precision}_c$ and $\mathrm{Recall}_c$ denote the precision and recall of class $c$, respectively. In our work, $C=8$, corresponding to the seven basic facial expressions and an additional category of $other$. Although our method is optimized with a cross-entropy-based objective function, all major comparisons in this paper are reported via the score of macro-averaged F1.

\subsection{Experimental Settings}
Our experiments follow the two-stage proposed method. In Stage~I, the DINOv2 ViT-L/14 is initialized from the official self-supervised pretrained checkpoint and finetuned on image-level FER datasets for visual adaptation with the proposed training strategies of PadAug and a training-only MoE head. We freeze the first eight transformer blocks and optimize the remaining layers for 12 epochs using AdamW with a backbone learning rate of $1\times10^{-5}$ and a head learning rate of $3\times10^{-4}$. Training techniques of mixup, label smoothing, and class-balanced supervision are applied, and the best checkpoint is selected according to the validation macro-averaged F1 score.

In Stage II, we construct a three-scale visual representation using face crops with scales of 0.9, 1.2, and 1.5 and average the DINOv2 features accordingly. Frame-aligned acoustic features are extracted from the pretrained Wav2Vec~2.0 large-lv60k ASR, using a centered temporal window of 0.50~s. A lightweight gated fusion head is trained on frame-level labels and the inference-time temporal smoothing is further evaluated with a fixed window size of 101 frames for temporal consistency. All experiments were conducted on the AutoDL platform using a single NVIDIA GeForce RTX 5090 GPU. Please review the setting details in the Appendix.

\subsection{Experimental Results}

\subsubsection{Validation Results}

Regarding performance, the best visual-only setting achieves a macro-averaged F1 score of 0.4344 on the official validation set. Introducing the module of audio-visual dual modal gated fusion improves the score to 0.5131 and yields a 5-fold mean score of 0.4842. After applying inference-time temporal smoothing, the system further reaches 0.5368 on the official validation set and 0.5122 $\pm$ 0.0277 under 5-fold cross-validation. These results validate our two-stage design of visual adaptation, dual-modal fusion and post-processing for time consistency.

\subsubsection{Test Result on ABAW Official Server}
In this paper, we also report the performance of our proposed method on the ABAW official challenge server. As shown in Table~\ref{tab:challenge_results}, our method achieves a total testing score of \textbf{0.391} on the official challenge server, ranking first in the EXPR challenge of ABAW 2026. This result further verifies the effectiveness of our method.

\subsection{Ablation Study}
To better investigate the contributions from individual modules and special techniques proposed in our method, ablation studies were conducted. Specifically, we examined the visual training techniques and the audio feature alignment, as well as the dual modal fusion method and the inference-time temporal smoothing. All the results are measured by the score of macro-averaged F1.

\subsubsection{Ablation Study on Visual Training Techniques}

The visual representation results from different training technique combinations are listed in Table~\ref{tab:visual_ablation}. 
Notably, first, re-cropping outperforms the official crop, since the latter is easier to lose facial details during image resizing. Second, model performance by averaging the three re-crop scales ($0.9/1.2/1.5$) surpasses that from each single scale, showing that the averaging operation can provide complementary information from individual cropped faces. Next, task-specific finetuning performs better than directly using the pre-trained DINOv2 model, indicating the necessity for model adaptation via extra data. Finally, attaching a MoE head on standard MLP brings additional gains from diversified learning of each expert.
In summary, combining the training strategy of PadAug and MoE, the facial images re-cropped and averaged in multi-scale, and trained via a finetuned DINOv2 model, returns the best macro-averaged F1 score of 0.4344.

\begin{table*}[t]
	\centering
	\caption{Comparison of different audio--visual fusion strategies. Results are reported as F1 on the official validation set and under 5-fold cross-validation.}
	\label{tab:fusion_ablation}
	\resizebox{\textwidth}{!}{
		\begin{tabular}{lcccccccc}
			\toprule
			Fusion & Params & Official Val F1 & 5-Fold Mean F1 & Fold-1 & Fold-2 & Fold-3 & Fold-4 & Fold-5 \\
			\midrule
			A only                & --    & 0.2901 & --     & --     & --     & --     & --     & --     \\
			V only                & --    & 0.4344 & --     & --     & --     & --     & --     & --     \\
			Concat + Linear       & 16K   & 0.4717 & 0.4527 & 0.4527 & 0.4203 & 0.4629 & 0.4946 & 0.4331 \\
			Concat + MLP          & 1.05M & 0.5079 & 0.4838 & 0.4670 & 0.4725 & 0.4983 & 0.5340 & 0.4474 \\
			Bilinear              & 1.32M & 0.4794 & 0.4502 & 0.4312 & 0.4341 & 0.4670 & 0.4714 & 0.4472 \\
			Dynamic Weighting     & 2.10M & 0.4712 & 0.4558 & 0.4466 & 0.4442 & 0.4651 & 0.4672 & 0.4557 \\
			Gated Fusion          & 2.37M & \textbf{0.5131} & \textbf{0.4842} & 0.4713 & 0.4783 & 0.4892 & 0.5235 & 0.4614 \\
			Cross-modal Attention & 6.31M & 0.5099 & 0.4837 & 0.4665 & 0.4702 & 0.4759 & 0.5112 & 0.4947 \\
			\bottomrule
		\end{tabular}
	}
\end{table*}

\subsubsection{Ablation Study on Audio Feature Alignment}

Table~\ref{tab:audio_alignment} evaluates the audio frame alignment strategies. The strategy of temporal window averaging (window mean) consistently outperforms the nearest-neighbor assignment, indicating short-term acoustic context is more informative than single-point alignment. A 0.50 s window achieves the best result of 0.2901, effectively capturing speech dynamics without excessive smoothing. We thus adopt this 0.50 s window mean strategy for our proposed method.

\subsubsection{Ablation Study on Dual Modal Fusion}

Table~\ref{tab:fusion_ablation} compares dual-modal fusion strategies. The audio--visual combinations consistently outperform the unimodal baselines, indicating that audio provides complementary cues for EXPR. Even a naive \textit{Concat + Linear} fusion improves the official validation F1 from 0.4344 to 0.4717. Among all methods in comparison, \textit{Gated Fusion} performs best, achieving an official validation F1 of 0.5131 and a 5-fold mean of 0.4842. Although \textit{Cross-modal Attention} is competitive, its much higher parameter count (6.31M vs.\ 2.37M) makes gated fusion a more economical choice, offering a favorable trade-off between performance and model complexity.

\subsubsection{Ablation Study on Temporal Smoothing}
\begin{figure}[t]
\centering
\includegraphics[width=0.95\linewidth]{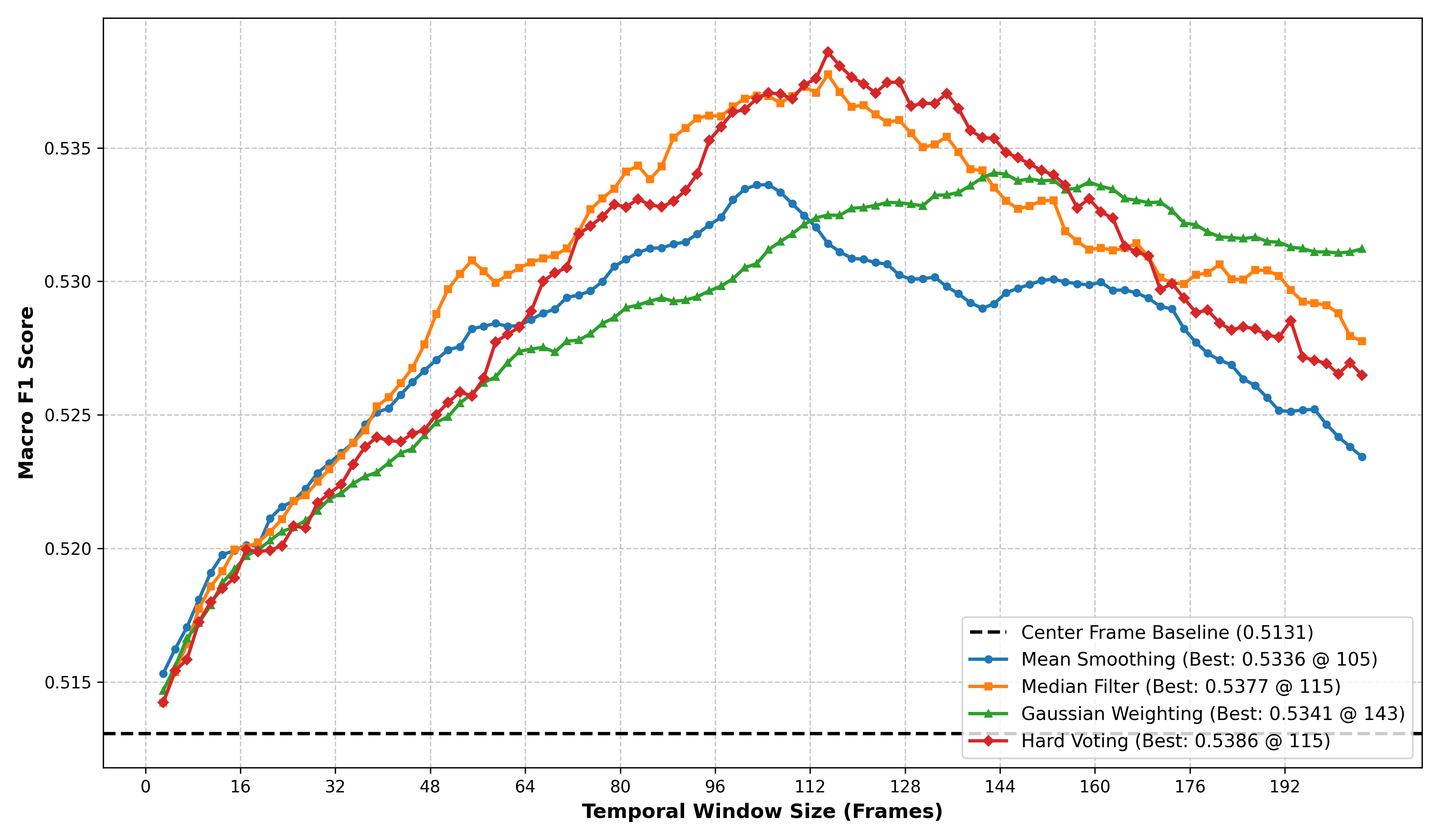}
\caption{Macro-averaged F1 score under different temporal smoothing strategies and window sizes.}
\label{fig:temporal_smoothing}
\end{figure}

We vary the temporal window from 3 to 205 frames with step size 2 and display the performance trends as well as the comparison results at a window size of 101 frames in Figure~\ref{fig:temporal_smoothing} and Table~\ref{tab:temporal_smoothing}, respectively. All strategies outperform the single-frame baseline, indicating that short-range temporal consistency helps stabilize predictions. The smoothing strategy of median filtering achieves the best results in our experiments, reaching 0.5368 on the validation set and 0.5122 $\pm$ 0.0277 under 5-fold cross-validation.

It should be emphasized that the 101-frame smoothing window is used since the performance from mean filtering starts to decline at 105 frames, indicating potential overfitting. Therefore, we mainly focus on the smoothing performances before the 105-frame setting in validation. During testing, the median filtering was further checked on multiple window sizes of 11, 51, 101, 151 and achieved the best score of \textbf{0.391} with 101-frame window. We therefore adopt median filtering with a 101-frame window in the final system.

\begin{table}[t]
\footnotesize
\centering
\setlength{\tabcolsep}{4.5pt}
\caption{Temporal smoothing results with a fixed window size of 101. Results are reported as F1 on the official validation set and under 5-fold cross-validation.}
\label{tab:temporal_smoothing}
\begin{tabular}{lcc}
\toprule
Strategy & Official Val F1 & 5-Fold Mean F1 $\pm$ Std \\
\midrule
Single-frame Baseline & 0.5131 & $0.4842 \pm 0.0212$ \\
Mean Smoothing        & 0.5335 & $0.5114 \pm 0.0303$ \\
Median Filter         & \textbf{0.5368} & \textbf{$0.5122 \pm 0.0277$} \\
Gaussian Weighting    & 0.5305 & $0.5098 \pm 0.0273$ \\
Hard Voting           & 0.5364 & $0.5095 \pm 0.0258$ \\
\bottomrule
\end{tabular}
\end{table}

\section{Conclusion and Future Work}
\subsection{Conclusion}

Our proposed method addresses the EXPR track of the 10th ABAW Challenge with a two-stage audio--visual framework for progressive expression recognition. In Stage~I, a pretrained DINOv2 ViT-L/14 backbone is fine-tuned on the AffectNet and RAF-DB datasets, together with PadAug and a training-only MoE head, to obtain a robust expression-aware visual encoder. In Stage~II, the adapted visual encoder is combined with frame-aligned Wav2Vec~2.0 acoustic features through a lightweight gated fusion module, followed by an inference-time temporal smoothing module, to obtain stable and temporally consistent predictions.

The proposed method achieves a macro-averaged F1 score of 0.5368 on the validation set, 0.5122$\pm$0.0277 under 5-fold cross-validation, and 0.391 on the official test set, ranking first in the EXPR track of ABAW~2026. Ablation studies validate the effectiveness of the key training strategies and functional modules in our framework. Experimental comparisons and ablation results further suggest that robust visual adaptation, lightweight fusion, and post-hoc temporal stabilization are more critical to strong performance than heavier or more sophisticated model designs in unconstrained affective and behavioral analysis.

\subsection{Future Work}

Our future work will focus on stronger visual modeling and more effective multimodal learning. In particular, adaptive face re-cropping, scale-aware feature aggregation, and finer-grained facial region modeling may further enhance the robustness of the visual branch, while lightweight audio adaptation and improved temporal aggregation may strengthen the contribution of the audio branch. Another promising direction is to extend the proposed framework to the joint learning of EXPR, valence-arousal, and action units, thereby enabling more general and robust affective representation learning in unconstrained environments.

{
    	\small
	\bibliographystyle{ieeetr}    
	\bibliography{main}
}

\clearpage

\appendix
\renewcommand{\thesection}{Appendix \Alph{section}}
\renewcommand{\thesubsection}{\Alph{section}.\arabic{subsection}}

\setcounter{figure}{0}
\renewcommand{\thefigure}{A\arabic{figure}}

\setcounter{table}{0}
\renewcommand{\thetable}{A\arabic{table}}

\setcounter{equation}{0}
\renewcommand{\theequation}{\arabic{equation}}
\setcounter{equation}{18}
\section{Training and Validation Details}

\begin{table*}[!t]
\centering
\small
\caption{Complexity statistics of the main modules used in our final system. FLOPs are reported under single-sample inference. For the visual encoder, the input is a face crop of size $224 \times 224$. For the audio encoder, the input is a 20-second waveform sampled at 16~kHz.}
\label{tab:complexity_stats}
\begin{tabular}{lcccc}
\toprule
Module & Input setting & Output shape & Parameters & FLOPs \\
\midrule
DINOv2 ViT-L/14 visual encoder
& face crop, $224 \times 224$
& $1 \times 1024$
& 303,227,904
& 81.437G \\

Wav2Vec 2.0 large-lv60k audio encoder
& 20-s audio, 16~kHz
& $999 \times 1024$
& 315,470,496
& 413.011G \\

Lightweight gated fusion head
& visual: $1 \times 1024$; audio: $1 \times 1024$
& $1 \times 8$
& 2,366,472
& 2.371M \\
\bottomrule
\end{tabular}
\end{table*}

\subsection{Training Details for Stage I}

Stage I adapts a pretrained DINOv2 ViT-L/14 backbone to facial expression recognition before frame-level audio--visual fusion. We initialize the visual encoder from the official self-supervised DINOv2 checkpoint. In our implementation, the patch embedding layer and the first eight transformer blocks are frozen, while the remaining blocks are fine-tuned. The model is trained for 12 epochs using AdamW with a backbone learning rate of $1\times10^{-5}$, a task-head learning rate of $3\times10^{-4}$, and a weight decay of $5\times10^{-2}$. We further apply gradient clipping with a maximum norm of 1.0 and use cosine annealing scheduling. To stabilize partial fine-tuning, we employ layer-wise learning-rate decay with a decay factor of 0.85 for the trainable transformer blocks, while bias and normalization parameters are exempted from weight decay.

Training images are augmented by a standard image-level pipeline consisting of random resized cropping to $224 \times 224$ with a scale range of $[0.8,1.0]$. We apply padding-aware augmentation (PadAug) during image training. With a probability of 0.6, we synthesize black padding artifacts on one or two image boundaries. The total padded area is capped at 20\% of the image area, with a minimum bar width ratio of 0.03 and a maximum shift ratio of 0.02 relative to the shorter image side. This augmentation is designed to mimic the boundary artifacts caused by large or imperfect face crops in raw videos.

We also apply mixup and label smoothing during the training in Stage I, where the mixup $\alpha$ is set to 0.2, which is enabled for all training batches, and the label smoothing factor is 0.1. The classification loss is implemented as class-weighted soft cross-entropy, where the soft targets come from the combination of one-hot smoothing and mixup. The best Stage-I checkpoint is selected according to the score of macro-averaged F1 on the validation set used for image-level visual adaptation.

The Stage-I classifier is a training-only gated MoE head attached to the 1024-dimensional CLS token of DINOv2. The routing branch of the MoE first compresses the 1024-dimensional input into a 256-dimensional hidden representation, followed by LayerNorm and GELU, and then predicts routing logits for four experts with a second linear layer. Each expert is a three-layer residual MLP stack based on SwiGLU blocks with expansion ratio 2.0. The expert outputs are fused by soft routing weights, followed by LayerNorm, dropout, and a final linear classifier. In the final Stage-I setting, the MoE head uses four experts, depth 3, dropout 0.6, and a maximum drop-path rate of 0.1. After Stage I, the MoE head is discarded, and only the adapted DINOv2 backbone is retained for Stage-II feature extraction.

\subsection{Training Details for Stage II}
Stage II starts with frame-level visual and audio features pre-extracted from Aff-Wild2, instead of processing raw images and waveforms in an end-to-end manner. For the visual branch, each target frame is re-cropped from the raw video at three face scales, namely 0.9, 1.2, and 1.5, and passed through the adapted DINOv2 encoder from Stage I. The resulting three 1024-dimensional visual features are averaged to form the final frame-level visual representation. For the audio branch, we use frame-aligned 1024-dimensional acoustic features extracted from a pretrained Wav2Vec 2.0 Large model. Only frames with valid EXPR labels are kept for training in Stage-II.

After extraction, each modality is optionally branch normalized and standardized by z-score normalization using mean and standard deviation computed on the training split only. In our implementation, both the visual and audio branches use feature-wise z-score normalization, with the training-set statistics saved and reused during validation. 

The optimized Stage-II uses the gated fusion head described in the main paper, with a hidden dimension of 512 and dropout 0.2. The fusion model in Stage-II is trained for 20 epochs using AdamW with a learning rate of $1\times10^{-4}$ and a weight decay of $1\times10^{-4}$. The batch size is set to 4096, and mixed-precision training is enabled. We use class-weighted cross-entropy as the training objective function in order to reduce the drawbacks from class imbalance. In addition, we apply element-wise input dropout to the fused input features during training, with dropout probabilities of 0.20 for the visual branch and 0.25 for the audio branch. The learning rate is scheduled by cosine annealing, and the best checkpoint is selected according to macro-averaged $F_1$ on the official validation split.

\subsection{Inference and Post-processing}

For validation and 5-fold cross-validation, inference is performed on frames that have valid image features and available labels. Frame-level logits are generated by the corresponding Stage-II model, and class-wise temporal smoothing is applied as a post-processing step. Final predictions are then obtained by taking the argmax of the smoothed logits. Different smoothing strategies and temporal window sizes are compared under this validation protocol.

For the test on the official server, we follow the official challenge requirement and generate predictions for all frames. To maintain dense frame-level outputs, when an exact frame-aligned feature is unavailable, the nearest available feature is used as a fallback. After obtaining the frame-level logits, we average the outputs of the five Stage-II models for 5-fold ensemble inference. Class-wise temporal median filtering is then applied to the ensembled logit sequence, and final predictions are obtained by taking the argmax of the smoothed logits. In the final system, the temporal median window is set to 101 frames.

\section{Computational Complexity}
\label{app:complexity}

As shown in Table~\ref{tab:complexity_stats}, the computational cost of the final system is mainly dominated by the pretrained visual and audio encoders, whereas the lightweight gated fusion head contributes only a marginal overhead in both parameters and FLOPs. The reported parameter counts and FLOPs are measured using fvcore. For the audio branch, the reported output shape corresponds to the hidden feature sequence extracted by Wav2Vec~2.0 from a 20-second audio chunk. In the final system, these acoustic features are further aggregated with a centered 0.50 s temporal window to produce frame-aligned audio representations for multimodal fusion.

\section{Details of the Gated Fusion Head}
\label{app:gate_details}

\begin{figure}[t]
	\centering
	\includegraphics[width=0.92\linewidth]{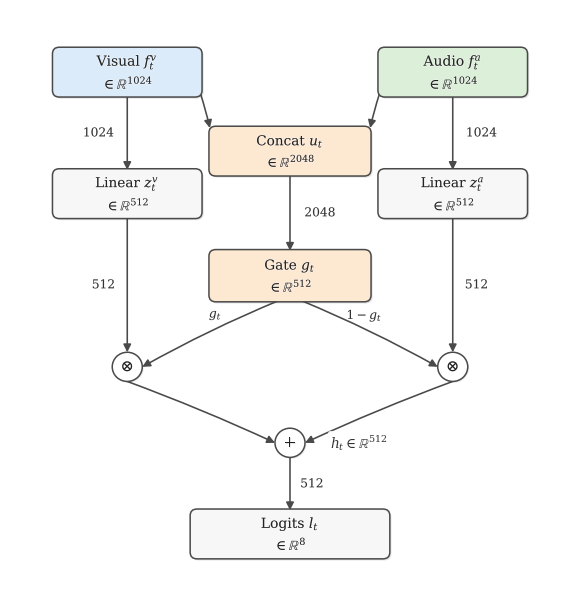}
	\caption{Detailed structure of the lightweight gated fusion head used in Stage II. The 1024-dimensional visual and audio features are first projected into a 512-dimensional space. A sigmoid gate is then predicted from the concatenated bimodal input to perform element-wise adaptive fusion, followed by a linear classifier for frame-level expression recognition.}
	\label{fig:gate_module}
\end{figure}

This section provides a detailed explanation of the gated fusion module used in our model. After the Stage-I adaptation process, the frame-level visual and audio features are denoted as $f_t^v \in \mathbb{R}^{1024}$ and $f_t^a \in \mathbb{R}^{1024}$, respectively, both represented as 1024-dimensional vectors.

As described in Section~3.2.3 and illustrated in Fig.~\ref{fig:gate_module}, the two modalities are first projected into a shared 512-dimensional space:
\begin{equation}
	z_t^v = W_v f_t^v + b_v,\qquad
	z_t^a = W_a f_t^a + b_a,
\end{equation}
where $W_v, W_a \in \mathbb{R}^{512 \times 1024}$ and $b_v, b_a \in \mathbb{R}^{512}$.

Next, the original bimodal features are concatenated as
\begin{equation}
	u_t = [f_t^v ; f_t^a] \in \mathbb{R}^{2048},
\end{equation}
and used to predict an element-wise gate:
\begin{equation}
	g_t = \sigma(W_g u_t + b_g),
\end{equation}
where $W_g \in \mathbb{R}^{512 \times 2048}$, $b_g \in \mathbb{R}^{512}$, and $\sigma(\cdot)$ is the sigmoid function. Therefore, each element of $g_t$ satisfies
\begin{equation}
	g_{t,i} \in (0,1), \qquad i=1,\dots,512.
\end{equation}

The final fused representation is obtained by element-wise interpolation between the projected visual and audio features:
\begin{equation}
	h_t = g_t \odot z_t^v + (1 - g_t) \odot z_t^a,
	\label{eq:gate_fusion}
\end{equation}
where $\odot$ denotes element-wise multiplication. Equivalently,
\begin{equation}
	h_t = z_t^a + g_t \odot (z_t^v - z_t^a).
\end{equation}
Hence, each dimension of $h_t$ is adaptively selected from the two modalities:
\begin{equation}
	h_{t,i} = g_{t,i} z_{t,i}^v + (1-g_{t,i}) z_{t,i}^a.
\end{equation}
This means that the fusion is performed independently in each feature dimension, rather than by introducing a dense cross-modal interaction matrix.

During training, dropout is applied to the fused representation, and the class logits are computed as
\begin{equation}
	l_t = W_o \,\mathrm{Dropout}(h_t, p) + b_o,
\end{equation}
where $p=0.2$ denotes a dropout rate of 20\%, $W_o \in \mathbb{R}^{8 \times 512}$, $b_o \in \mathbb{R}^{8}$, and $l_t \in \mathbb{R}^{8}$ corresponds to the logits over the eight expression classes.

\paragraph{Gradient interpretation.}
Let $\mathcal{L}$ denote the Stage-II training loss $\mathcal{L}_{\text{stage2}}$ defined in Eq.~(13) of Section~3.2.4, and let $\delta_t = \partial \mathcal{L}/\partial h_t$. From Eq.~\eqref{eq:gate_fusion}, the gradients with respect to the two projected modalities and the gate are
\begin{equation}
	\frac{\partial \mathcal{L}}{\partial z_t^v}
	= \delta_t \odot g_t,
	\qquad
	\frac{\partial \mathcal{L}}{\partial z_t^a}
	= \delta_t \odot (1-g_t),
\end{equation}
and
\begin{equation}
	\frac{\partial \mathcal{L}}{\partial g_t}
	= \delta_t \odot (z_t^v - z_t^a).
\end{equation}
Therefore, the gradient received by each modality is explicitly modulated by the gate value. A larger $g_{t,i}$ assigns more learning signal to the visual branch in the $i$-th dimension, while a smaller $g_{t,i}$ assigns more learning signal to the audio branch. Moreover, all else being equal, the gate receives a stronger update when the discrepancy $|z_{t,i}^v - z_{t,i}^a|$ is larger, which makes the fusion more sensitive to dimensions where the two modalities provide different evidence.

Since $g_t$ is generated by a sigmoid layer, we have $g_t = \sigma(W_g u_t + b_g)$, and the gradient with respect to the corresponding pre-sigmoid activations is further modulated by the sigmoid derivative term $g_t \odot (1-g_t)$. This provides a simple and stable optimization path for the gate while keeping the fusion head lightweight.

\paragraph{Why gating instead of attention.}
In our setting, both modalities have already been encoded into high-level semantic representations by strong pretrained encoders. The role of the fusion head is therefore not to perform heavy cross-modal token matching, but to conduct lightweight adaptive selection and combination. Compared with attention-based fusion, the proposed gate provides a more explicit inductive bias and achieves greater parameter efficiency, consistent with the comparison in Table~4 of Section~4.4.3. In addition, the gated fusion head itself has a low computational overhead, as indicated by its small FLOP contribution reported in the appendix. This design is therefore well suited to our objective of simple and effective multimodal fusion.

\section{Summary of Minor Revisions in the Camera-Ready Version}
\label{app:revisions}

For completeness, we summarize the minor issues raised during review and the corresponding revisions incorporated into the camera-ready version. These changes improve clarity, formatting compliance, and reference consistency, but do not affect the proposed method or the reported experimental results.

\noindent\textbf{Comment 1:} The caption of Fig.~1 should describe the method in more detail and be sufficiently self-contained.

\noindent\textbf{Revision made:} We rewrote the caption of Fig.~1 to provide a clearer and more self-contained summary of the proposed framework. The revised caption now explicitly describes the multi-scale facial crops, the frame-aligned audio representation, the gated fusion module, and the inference-time temporal smoothing step, so that the overall method can be understood more easily from the figure itself.

\medskip
\noindent\textbf{Comment 2:} A formatting issue was noted in the review version.

\noindent\textbf{Revision made:} We corrected the manuscript formatting in the camera-ready version and aligned the paper with the required final template. The line-number placement issue observed in the review version is therefore no longer present.

\medskip
\noindent\textbf{Comment 3:} The references to the current ABAW competition and its corpora should follow the organisers' guideline.

\noindent\textbf{Revision made:} We updated the reference list in accordance with the organisers' guideline for citing the current ABAW competition and its corpora. Citations not required by the guideline were removed. At the same time, we retained the key official references that are necessary for introducing the EXPR track and the overall multi-task challenge context, so that the background remains both compliant and sufficiently informative.

\end{document}